\documentclass[format=acmlarge,balance=false]{acmart}
\author{
  Ravit Sharma, Wojciech Romaszkan, Feiqian Zhu, Puneet Gupta, Ankur Mehta
}
\usepackage{pgfplots}
\AtBeginDocument{%
  }

\begin{document}

%\title{Enabling Accessible Machine Learning: Exploring Model Compression for Microcontrollers using Multi-Modal Biometric Data}
\title{Cost-Driven Hardware-Software Co-Optimization of Machine Learning Pipelines}

\begin{abstract}
Researchers have long touted a vision of the future enabled by a proliferation of internet-of-things devices, including smart sensors, homes, and cities. Increasingly, embedding intelligence in such devices involves the use of deep neural networks. However, their storage and processing requirements make them prohibitive for cheap, off-the-shelf platforms. Overcoming those requirements is necessary for enabling widely-applicable smart devices. While many ways of making models smaller and more efficient have been developed, there is a lack of understanding of which ones are best suited for particular scenarios. More importantly for edge platforms, those choices cannot be analyzed in isolation from cost and user experience. In this work, we holistically explore how quantization, model scaling, and multi-modality interact with system components such as memory, sensors, and processors. We perform this hardware/software co-design from the cost, latency, and user-experience perspective, and develop a set of guidelines for optimal system design and model deployment for the most cost-constrained platforms. We demonstrate our approach using an end-to-end, on-device, biometric user authentication system using a \$20 ESP-EYE board.
\end{abstract}


\maketitle
%\todotbd{General todos}
%\todors{Ravit todo}
%\todowr{Wojciech todo}

%%%%%%%%%%%%%%%%%%%%%%%%%
% Introduction
%%%%%%%%%%%%%%%%%%%%%%%%%
\section{Introduction} \label{sec:uist23:intro}

The combination of Internet-of-Things (IoT) and Deep Learning (DL) trends has created an enormous demand for ultra-low footprint machine learning models, commonly referred to as TinyML \cite{Dutta2021TinymlSurvey}. On-device, or near-sensor, inference ensures privacy while avoiding high energy and latency cost of offloading computation to the cloud \cite{Banbury21Mlperftiny}. Enabling more complex algorithms on low-cost, microcontroller-based systems has the potential of \textit{making access to smart devices ubiquitous}. Broad availability, low cost, and ease-of-use would, in turn, make it possible for people to experiment with an increasingly-broader range of applications that improve human-computer interaction, such as audio and visual wake words, context recognition, and user verification \cite{Dutta2021TinymlSurvey}. However, achieving it requires unprecedented efforts on \textit{co-optimization} of algorithms and hardware to make large and computationally complex models usable on devices with very limited memory and processing power \cite{Romaszkan20203pxnet}. \par

Multiple techniques have been proposed to address model compression: quantization \cite{Abadi16TensorFlow, Lin16FXPQuant, Gupta2015NnQuant16bit, Romaszkan20203pxnet}, which uses lower precision numbers for more efficient storage and computation, pruning \cite{Cun1990NnBrainDamage, Han2015NnPruning, Romaszkan20203pxnet} which removes inconsequential weights, compressed models \cite{Iandola2016SqueezeNet, Howard2017NnMobileNet}, and optimized software libraries \cite{Lai2018MlCmsis, David21TFLiteMicro}. While all the above \textit{knobs} are readily available to machine learning researchers, it is not obvious how they interact with hardware configurations, given the specific set of constraints, e.g., \textit{cost, latency, size, and user experience}. While approaches like neural architecture search (NAS) can automate finding feasible solutions, they are often targeted at larger models \cite{Dai2019ChamNet}, are constrained in scope \cite{Lin20McuNet, saha_tinyodom_2022}, and rarely optimize the cost of the overall system. As a result, deploying efficient ML models on edge devices in a cost-aware fashion currently requires significant expertise, which makes them inaccessible to a vast pool of potential developers. \par

% \todotbd{Remove pruning if not using.}

Our goal is to make \textit{ML systems both cheap and easy to deploy}. To this end, we empirically explore and co-optimize the interaction of various compression techniques and component choices, from a cost-centric perspective, to establish \textit{a set of guidelines} for deploying ultra-compact models on memory- and compute-constrained devices. We evaluate how varying quantization, model architecture, and multi-modality affects accuracy, and develop a set of optimal choices given cost, size, and latency constraints. Our findings make it possible to easily deploy complex models on cheap, off-the-shelf devices. We demonstrate this using a robust biometric authentication application deployed on a low-cost microcontroller board equipped with a camera and microphone module. \par

% \todotbd{Adjust as necessary}
The contributions of this work are as follows:
\begin{itemize}
    \item We explore how different model compression techniques and multi-modality affect hardware requirements when deploying compact models ($<$10MB, $<$1MB).
    \item Based on the above, we develop a set of guidelines for choosing an optimal compression strategy given cost, size, and latency constraints.
    \item We demonstrate the utility of our approach by building a robust, low-cost, biometric user authentication platform.
\end{itemize}

%%%%%%%%%%%%%%%%%%%%%%%%%
% Motivation
%%%%%%%%%%%%%%%%%%%%%%%%%
\section{Accessibility of Edge ML} \label{sec:uist23:motiv}

Edge machine learning inference has been extensively explored in recent years as a way to lower its cost, improve privacy, and remove the communication energy and latency involved in offloading computation to the cloud \cite{Dutta2021TinymlSurvey}. Various model compression techniques have been used to make such applications feasible on severely constrained microcontroller-class devices. Compressed neural network models like SqueezeNet \cite{Iandola2016SqueezeNet} and MobileNet \cite{Howard2017NnMobileNet} have been developed, specifically targeting low memory footprints. Dedicated runtime libraries have also been developed, like ARM CMSIS-NN \cite{Lai2018MlCmsis}, TensorFlow Lite \cite{David21TFLiteMicro}, or  Microsoft EdgeML \cite{Kumar2017NnResourceEfficient, Gupta2017NnProtoNn}. Pruning networks to high sparsity levels to save storage and computation has also been extensively explored \cite{liberis_differentiable_2022, Romaszkan20203pxnet}. \par

One approach that has received significant attention is quantization, which allows for neural network inference to use 16- or even 8-bit precision \cite{Gupta2015NnQuant16bit, Lin16FXPQuant, David21TFLiteMicro}. The benefits of using quantization are twofold: it requires less storage and makes it possible to use cheaper compute. Those benefits have been translated to significant performance improvements \cite{Lai2018MlCmsis, Yao2020NnHawqV3}. The 8-bit precision inference is a de facto standard on embedded systems since it can frequently match floating-point accuracy and is the lowest precision natively supported in computation \cite{Lin20McuNet, David21TFLiteMicro}. Further, microcontrollers often may not contain floating-point units (FPUs), making floating-point computation prohibitively expensive \cite{Lai2018MlCmsis}. \par

Even the most extreme form of quantization---binarization, which restricts all values to +1 and -1---has been demonstrated to work on certain models \cite{Courbariaux2015NnQuantBinary}. Besides storage compression, binarization offers impressive performance improvements by replacing integer multiplication with bitwise XNOR operations \cite{Rastegari2016NnXnornet,Courbariaux2016NnBinarized,Hubara2016}. Due to their efficiency, multiple software \cite{Yang2017NnBMXNet, Hu2018NnBitFlow, McDanel2017NnBinEmbedded, Pedersoli2018NnEspresso}, and hardware \cite{Umuroglu2017NnFinnFpga, Ando2017NnBrein, Li2017BinNnFpgaAccel, Jiang2017NnXnorPop, Conti2018, Bahou2018, Fraser2017, Bankman2019} implementations of binarized neural networks have been proposed; as well as further optimizations, such as combining binarization with pruning \cite{Romaszkan20203pxnet}, or memoization \cite{Liu21TcpNet}. However, BNNs come with disadvantages, the most important of which is accuracy degradation \cite{Elhoushi2019NnDeepShift, Fromm2019BinNnRiptide}. To address this issue, one of the early efforts, DoReFa-Net, proposed increasing the bitwidth of activations and weights while retaining efficient XNOR-based implementations \cite{Zhou2016NnQuantDorefa}. Multiple following works have expanded on this idea of \textit{multi-bit networks} to make high-accuracy BNNs possible \cite{Elhoushi2019NnDeepShift, Fromm2019BinNnRiptide, Zhang2021BitlineShift}. \par

Clearly, there is no shortage of methods to \textit{make models more compact}, but how does one choose the best one, given specific application constraints? There is a general lack of guidelines in that aspect. Neural architecture search approaches, which automate the process of finding an optimal model, offer some answers \cite{Lin20McuNet, Dai2019ChamNet, saha_tinyodom_2022}. However, they often target much larger models \cite{Dai2019ChamNet, Ding20AutoSpeech}, or are constrained to only explore one aspect of compression, like model architecture \cite{Lin20McuNet}, or quantization level \cite{Yao2020NnHawqV3}, rarely the overall system cost. Because of that, the field of edge machine learning is currently limited to a small group of experts who have the knowledge, time, and resources to experiment with the above knobs. There is a need for a more nuanced approach - one that takes into account multiple different compression schemes given cost, size, and latency constraints, and evaluates how they interact when trying to deploy models with a size in single MBs.  \par

%%%%%%%%%%%%%%%%%%%%%%%%%
% Prior Work
%%%%%%%%%%%%%%%%%%%%%%%%%
\section{Authentication for Accessibility} \label{sec:uist23:prior}

User authentication is a crucial application in human-computer interaction, as it can be used to restrict access to critical resources and personal data \cite{Kataria13SurveyBiometric, Barra19BiometricData}. Enabling low-cost and easy-to-use authentication is therefore crucial for the broader accessibility of edge systems. Among various authentication methods, biometric ones, which use measurable and immutable physiological or behavioral characteristics of individuals, are especially popular \cite{Rui19SurveyBiometric}. They are intuitive to use, do not require memorization of, e.g., passwords, and the difficulty in reproducing biometric stimuli provides more robustness to attacks \cite{Rui19SurveyBiometric}. Increasingly, deep learning is used for biometric authentication due to improved accuracy compared to prior methods \cite{Minaee21BiometricSurvey, An22FaceRecognition}. However, most deep learning models are not real-time or memory-efficient, especially on low-cost devices \cite{Minaee21BiometricSurvey}, resulting in a significant price tag for existing authentication systems. Offloading the computation to the cloud may not be possible due to the importance of privacy of biometric data \cite{Rui19SurveyBiometric, Ali18MultimodalAuth}. It also introduces an additional layer of complexity needed when designing such systems, increasing the expertise level required. Developing highly-compressed, cost-aware, robust deep learning approaches to biometric authentication is, therefore, a necessary step towards enabling secure, affordable, and accessible computing systems. \par

Biometric authentication has been demonstrated on a host of characteristics, such as face, iris, fingerprint, voice, breathing, or keystroke dynamics \cite{Rui19SurveyBiometric, Ali18MultimodalAuth, chauhan_performance_2018}. Among those, face and voice recognition exhibit a set of advantages that make them a popular choice. They are both highly universal, meaning that they are applicable to the majority of users, they can be obtained using widely available devices such as cameras and microphones, and they are broadly accepted by users as authentication methods. Voice, in particular, has high uniqueness and permanence, meaning it is easy to distinguish individuals and does not change significantly with time \cite{Rui19SurveyBiometric}. Both face and voice authentication has been widely studied, and various authentication algorithms have been proposed for both \cite{Rui19SurveyBiometric}. \par

While prior works have shown edge biometric authentication systems \cite{Giorgi19VoiceRecognition, Kunik17RpiIrisRecognition, Shaout21SmartDoorbell}, they have done so while ignoring the cost of such devices. They employ expensive FPGA devices \cite{Giorgi19VoiceRecognition} or single-board computers \cite{Shaout21SmartDoorbell, Kunik17RpiIrisRecognition} that can cost hundreds of dollars, making them prohibitive to many enthusiasts. We aim to enable such applications on systems costing single or lower-double-digits dollars. Others sidestep the issue by processing data in the cloud \cite{Alvarez15SmartDoorbell, Joshi20SmartDoorbell}, which involves privacy risks, additional complexity, and increased latency caused by network communication \cite{Dutta2021TinymlSurvey}. We want to make our system more secure and predictable by keeping all the data and processing entirely on the device. \par 

Prior works have shown the importance of multi-modality and fusion in biometric authentication, as it can improve accuracy and resilience to attacks \cite{Stylios21BehavBiometric, Rui19SurveyBiometric, Kataria13SurveyBiometric, Minaee21BiometricSurvey, Ali18MultimodalAuth, Chao16MultimodalBiometric}. However, many of those works ignore the computational and storage cost of implementing multi-modality and ignore the pertinent \textit{edge-only} deployment scenario \cite{Chen20VoxCelebMultimodal, Chao16MultimodalBiometric, Ali18MultimodalAuth}. Naively, including additional modalities, involves additional storage and processing, not to mention the cost of sensors used to acquire the data. However, prior works have shown that employing multi-modality can result in better accuracy for a smaller size, compared to models using only a single modality \cite{saha_tinyodom_2022}. This means that we can utilize multi-modality as an additional knob in our exploration of machine learning compression. However, the cost trade-off between potentially smaller memory and additional sensors needs to be carefully considered. \par

To summarize, there is a pressing need for a better understanding of how ultra-compact neural network models should be chosen and deployed. In the remainder of this paper, we show how this understanding can be developed and utilized towards biometric user authentication as an example application that can greatly benefit it, enabling cheaper, more secure embedded systems. \par

%%%%%%%%%%%%%%%%%%%%%%%%%
% Ubiquitous ML
%%%%%%%%%%%%%%%%%%%%%%%%%
\section{Constructing a Microcontroller Search Space}

A variety of factors contribute to the cost of an IoT system, including the chosen processor, sensors, the amount of memory (PSRAM), and storage (Flash) needed. These factors affect the minimum cost of a board on which the model can run. We use information about each of these determining factors to generate a model search space over which we will search for the best performance-cost trade-off.\par

\subsection{Neural Network Execution Runtime}

% Here describe your execution flow: weights + code in Flash; activations in PSRAM; dual-core means two models run in parallel. Explain that flash size is dictated my model size while PSRAM by activations. Say that more complex memory mappings and dataflows may be possible but your evaluations assume the above.

Our setup for running a small ML model on a microcontroller board is as follows. We store the model's weights and model execution code, including preprocessing code, in nonvolatile storage (i.e. flash). By doing so, we ensure that the board does not lose its functionality after being powered off. Further, this read-only data is stored in cheaper Flash storage instead of expensice fast memories (e.g., PSRAM). Intermediate activations calculated during the execution of the model are stored in memory (e.g., PSRAM). Although more complex mappings of weights and activations onto flash/PSRAM are possible depending on the choice of dataflows,  we limit ourselves to this fairly common setup.

\subsection{Hardware Options}

From a hardware perspective, the parameter space includes storage size, memory size, sensors, and processor cost.

\subsubsection{Nonvolatile Storage Size}

The program code size plus the parameter size, which is the total size occupied by the weights of the model, must strictly be smaller than the nonvolatile storage capacity of the microcontroller (usually Flash memory). For example, if the parameter size of the model is 3MB and the program size is 0.25MB, a minimum Flash size of 3.25MB is required. In practice, the size of off-chip flash is manufactured in powers of two (e.g. 1MB, 2MB, 4MB), so we would round up to the next power of two when deciding flash size. The parameter size of a model is affected by whether the model is unimodal or multimodal, the complexity of model architecture, and the quantization scheme used. Because a multimodal model is composed of smaller subnetworks, it usually has a larger parameter size \cite{radu_multimodal_2018}. If a more complex model is used, it likely has more weights and thus a larger parameter size \cite{He2015NnResnet}. Finally, the quantization scheme, which includes the precision of the weights and activations, affects the parameter size \cite{Romaszkan20203pxnet}. \par

In our use case, we analyze models of three precisions: floating-point (32-bit), fixed-point (8-bit), and XNOR (1-bit to 3-bit, bitwidth may be different for weights and activations) \cite{Romaszkan20203pxnet, Zhou2016NnQuantDorefa}. We tested three combinations empirically found to generate the best accuracy: 3/1 (3-bit activations, 1-bit weights), 2/1, and 2/2. To restrict the search space, we do not mix and match quantization schemes within each model; all layers within any given model have the same quantization scheme. \par

\subsubsection{Memory Size}

In addition to storage, random-access memory (RAM) is necessary for storing inputs and intermediate results. For simplicity, we compute the minimum memory size as the maximum activation size over all layers of the computational graph. While techniques such as tiling can lower this requirement, we omit them to further constrain our search space \cite{Cowan2018}. As with Flash size, the minimum required memory size is affected by the model's modality, architecture, and quantization scheme. \par

Results in prior works indicate that ResNet architectures tend to perform well for face and voice authentication \cite{Ding20AutoSpeech}. %TODO: Citation for this claim
Hence, we restrict our search space to small ResNet-style models whose number of parameters varies from under half a million parameters to just over ten million parameters. Specifics of the models tested are discussed further in the Section \ref{sec:uist23:meth}.

\subsubsection{Sensors}

The third factor affecting the cost of a biometric device is its input modalities. We focus on two commonly used ones: face and voice recognition. Therefore, either a microphone, a camera sensor, or both in the multi-modal system, are required on the microcontroller board. The price of sensor models varies depending on the resolution of the sensor, like the resolution of the camera, or the sampling rate of the microphone). Because the pre-processing pipeline and input size for a model is usually fixed, the training and model configuration dictates the required specifications of the sensor. Nevertheless, sensors of lower quality can be used, e.g., lower resolution cameras, though usually at the expense of degraded accuracy \cite{zou_very_2014}.

\subsubsection{Processor}

Lastly, the cost of the microcontroller is affected by its processor. Although multiple factors determine the price of a processor, the main distinction that we will focus on in this study is single- vs. multi-core. This selection is independent of the model's modality choice, as any model can, in theory, be implementet as single- or multi-threaded to take advantage of multiple cores.. However, it is more complicated to split the workload of a single-modality system over multiple cores than it is for a multimodal one. This is because the workload on a multimodal system, especially one with independent processing pipelines, can be easily parallelized, which creates "free" multi-threading opportunities. For example, a dual-core processor can run a face/voice multimodal model in half the time of a single-core processor (assuming the architectures of the two networks are identical). More generally, multicore processors are advantageous (particularly for multimodal models), though the scheduling dynamics will vary. \par

\subsection{Machine Learning Model Options}

From a software and ML development perspective, model architectural choices can interplay with the model's cost, which is advantageous for multimodal models with influencing factors.

\subsubsection{Model complexity}
The complexity of the model, summarized by its parameter count and number of layers, is an indication of its expressive power. Assuming standard architectural practices are followed, such as in a typical convolutional neural network (CNN), increasing the complexity of a model raises its expressive power, which, in turn, increases its ability to produce more accurate results. Increasing model complexity is a useful means to improve its performance, but as will be illustrated by our results, it tends to yield diminishing returns.

\subsubsection{Quantization}

The decision of an ML developer to quantize an ML model is another factor influencing the capabilities of a neural network. By discretizing the weights and activations, the number of distinct states that the model can occupy is reduced. If quantization is applied, more aggressive quantization, i.e., fewer bits per value, further increases the  ability of the model to converge well.

Quantization is usually applied to decrease the nonvolatile storage space consumed by the weights of a neural network and increase inference speed. However, these advantages come at the expense of decreased accuracy.

\subsubsection{Multimodality}

Lastly, varying the input modalities of a model is a tactic that can help improve a model's performance by enabling a user to rely not just upon one but multiple streams of data. This decision makes the model more resistant to anomalies in one input stream of data by supplementing that input with another data source. For example, camera and microphone inputs may be used in combination to authenticate an individual. Several methods of enabling multimodality in a model exist, like feature concatenation, or ensemble classifiers \cite{radu_multimodal_2018}. Of course, the possibility of adding modalities is constrained by the dataset(s) and use-case. For example, adding an IMU-sensor input to an image-based plant leaf classifier would not be practical, since there is no correlation between the stability of the camera and the species of plant. However, adding acceleration input to an image-based predictive maintenance application on a factory line is useful because abnormal vibrations may correspond to an anomaly on the factory line.

\begin{comment}
Finally, the degree of independence between the modalities used is associated with better results, since the two data streams capture related but distinct sources of information. For example, replacing a single camera with a stereo camera in an authentication application may enable depth perception and make the application more occlusion resistant. However, in comparison to a model with input from a single camera and a single microphone, the model with input from two cameras falls short because an audio sample would yield much more novel information about the person than a second camera would. Nonetheless, in authentication, there is an important correlation between camera and audio inputs, which will be further discussed in Section \ref{ssec:uist23:disc:modal}.
\end{comment}

\subsection{Interrelation Between Hardware and ML Model Choices}

The hardware choices and ML-related choices made by a TinyML developer are constrained by one another. For example, having a multimodal model with inputs from a LiDaR and Microphone sensor necessitates having the relevant sensors on the board. As another example, the parameter size of a model (which takes model complexity and the quantization scheme into account) must not exceed the available nonvolatile storage on the board. Table \ref{tab:uist23:meth:correlation} summarizes these results by indicating the degree of correlation between the hardware and ML-related selection criteria.

\begin{comment}

\begin{table}[htbp]
\caption{Correlations between factors influencing hardware and ML models. ++ indicates a strong correlation, + indicates 
a weak correlation and a blank indicates no correlation.}
\begin{tabular}{lllll}
\toprule
& Model Complexity & Weight Quantization & Activation Quantization & Multimodality \\ 
\midrule
Nonvolatile Storage Size & ++ & + & ++ & + \\ 
Memory Size & + & ++ & ++ & + \\ 
Sensors & + & + & & ++ \\ 
Processor/Cores & & & & + \\
\bottomrule
\end{tabular}
\label{tab:uist23:meth:correlation}
\end{table}

\end{comment}
\begin{table}[htbp]
\caption{Correlations between factors influencing hardware and ML models. ++ indicates a strong correlation, + indicates 
a weak correlation, and a blank indicates no correlation.}
\begin{tabular}{llll}
\toprule
& Model Complexity & Quantization & Multimodality \\ 
\midrule
Nonvolatile Storage Size & ++ & ++ & + \\ 
Memory Size & ++ & ++ & + \\ 
Sensors & + & & ++ \\ 
Processor/Cores & & & + \\
\bottomrule
\end{tabular}
\label{tab:uist23:meth:correlation}
\end{table}

The constraints indicated above are summarized in the following bulleted list:
\begin{itemize}
    \item Model complexity strongly correlates with nonvolatile storage and memory size since a complex model has a higher parameter count, consumes more storage/weight, and has a higher peak memory size. Complexity weakly correlates with the number of sensors because a multimodal model is usually larger to process multiple forms of input.
    \item Quantization strongly influences both the minimal nonvolatile storage space and memory size of hardware because quantizing activations and weights reduces the minimum peak activation size and parameter storage size, respectively.
    \item Multimodality has a one-to-one correspondence with the number of sensors needed to run the model. Additionally, due to implications for model complexity and parallelism, it is weakly correlated with nonvolatile storage size, memory size, and the number of cores in the processor chosen.
\end{itemize}

%%%%%%%%%%%%%%%%%%%%%%%%%
% Methodology
%%%%%%%%%%%%%%%%%%%%%%%%%
\section{A User Authentication Case Study} \label{sec:uist23:meth}

%\todors{Implementation specifics: platforms, models, libraries, training setup, optimizations, deployment, measurement - how did we get the results, and how could other people reproduce them.}

%\subsection{Hardware and ML-Model Choices in Authentication}
As introduced in Section \ref{sec:uist23:prior}, we explore the interplay between hardware and ML-Model choices through the use case of face/voice authentication. In order to accomplish this, we construct our model search space by testing several models of various complexity, applying three quantization schemes of varying degree, and combining the face and speaker inputs to create a multimodal model. Due to the strong interrelatedness between hardware and ML-model factors, we first vary factors pertaining to ML-model, leading to a range of values across the hardware factors as well.

\subsection{Model Complexity}
ResNet is a modular architecture of Convlutional Neural Networks (CNNs) first proposed in \cite{He2015NnResnet}. In past works, ResNet models have been shown to perform well on audio-visual inputs \cite{Chen20VoxCelebMultimodal}, indicating their suitability for authentication. Additionally, ResNet models are highly modular as they are composed of repeating units of identity and convolutional blocks. This property enables us to easily vary the complexity of our ResNet model in a controlled fashion. Because the smallest "standard" ResNet model, ResNet18, is already on the larger side for our desired model complexity, we further downsize ResNet18 by reducing the number of identity and convolutional blocks within it. We vary the sequence of blocks in the ResNet model, which corresponds to the "layers" constructor variable in the TorchVision implementation of ResNet \cite{torchvision_resnet}. This reduction allows us to limit our testing to a small family of ResNet networks with parameter counts ranging from 380K to 11.5M. The range of ResNet models over which we test is described in Table \ref{tab:uist23:meth:models} \cite{He2015NnResnet}.

In this table, "blocks" is a list specifying the sequence of layers used to construct a particular model. Each element in the list indicates a number of layers after which the number of channels is doubled, and the input dimensions are halved. For example, [2, 2, 2], corresponding to ResNet-14, indicates that the model should contain a total of 6 blocks, with the third and fifth blocks being convolutional blocks that double the number of channels and half the width/height of the input activation. However, rather than the implementation details, we emphasize that the "blocks" variable provides us with a controlled mechanism to vary the complexity of our authentication model. \par

\begin{table}[htbp]
\caption{List of custom ResNet models we experimented with and their floating-point parameter size.}
\begin{tabular}{lll}
\toprule
Model Name & Blocks & Float Param. Size [MB] \\ 
\midrule
ResNet-6 & [1, 1] & 1.453 \\ 
ResNet-8A & [2, 1] & 1.736 \\ 
ResNet-8B & [1, 2] & 2.583 \\ 
ResNet-10 & [2, 2] & 2.860 \\ 
ResNet-14 & [2, 2, 2] & 11.124 \\ 
ResNet-18 & [2, 2, 2, 2] & 43.564 \\
\bottomrule
\end{tabular}
\label{tab:uist23:meth:models}
\end{table}

\subsection{Quantization and Sparsity}

We use Quantization-Aware Training (QAT) \cite{Rusci2019NnQuant} to train models of two quantization schemes: fixed-point and XNOR. To do so, we make the following modifications to the ResNet architecture. Firstly, we replace all floating-point Conv/FC layers within each ResNet block with their multibit XNOR or fixed-point equivalents. Throughout this paper, we use XNOR A/B to denote an XNOR-quantized network with A-bit activations and B-bit weights. For example, in an XNOR 2/1 ResNet18, we will replace each floating-point Conv2d layer with an XNOR Conv2d layer that binarizes the input activations and weights before computing the output. Secondly, for compatibility with our C implementation, we change the stride of the first layer from 2 to 1 and instead double the kernel size of the subsequent MaxPool2d layer from 2 to 4. Thirdly, we place the first max pooling immediately after the first convolutional layer instead of after the batch normalization and activation (ReLU). Finally, we normalize the output embedding to have a Euclidean norm of 1. All of the above are applied to all precision points for consistent evaluation. \par

\subsection{Training Setup \& Metrics} \label{ssec:uist23:meth:train}

\subsubsection{Input preprocessing}

The input of a face model is a 224x224 RGB image with a centered headshot, and its output is a 512x1 embedding, representing the learned position of the face in 512-dimensional hyperspace. The speaker mode takes a spectrogram of a 3-second audio clip as input, and outputs a 512x1 embedding, like the face model. Given two individually trained face and speaker models, we are able to construct a multimodal "fusion" network with both modalities with no additional training by simply concatenating their embeddings. For each model type, the embedding is then fed to a classification layer of size 512x5994, where 5994 is the number of unique identities in the training dataset. Because both the face and speaker models take similarly sized 3D tensors as input, we are able to use identical architectures to train both. The fusion model is not trained separately. Instead, it simply concatenates the output of the pretrained face and speaker models, outputting a 1024x1 embedding. \par

\subsubsection{Training setup}

We train our models in Pytorch (version 1.9.0), using cross-entropy loss on the VoxCeleb dataset, which contains images and voice samples from thousands of individuals \cite{chung_voxceleb2_2018}. The quantization and binarization support is based on open-source code from \cite{Romaszkan20203pxnet}. The source code was adjusted to allow multi-bit processing to enable accuracy-runtime trade-offs, similarly to \cite{Fromm2019BinNnRiptide}. \par
    
In terms of numerical precision, we use floating-point as a baseline, 8-bit fixed-point, and multi-bit XNOR networks with 1, 2, and 3 bits used for activations, and 1-2 bit weights. We choose not to extend binarized precision beyond 3 bits, as we have determined that doing so does not improve runtime over 8-bit fixed point networks, due to overheads involved in packing, unpacking, and non-vectorized bitcount operations \cite{Romaszkan20203pxnet}. \par

\subsubsection{Model performance}

To compare model performance we use the equal error rate (EER) metric \cite{Rui19SurveyBiometric}. EER refers to the point at which false rejection and false acceptance rates are equal. The lower the EER, the higher is the accuracy of the authentication system \cite{Rui19SurveyBiometric}. EER is a useful metric for authentication because it allows us to evaluate the performance of our model on identities that have not been seen before. To evaluate a model on a new identity, a new embedding must be registered corresponding to the identity by running the face and/or voice sample through the model and storing the result. Subsequently, the face and/or voice embedding of a new user that presents him or herself to the camera will be calculated and compared to the reference embeddings calculated earlier. The user will be said to be accepted and authenticated if the Euclidean distance is below a threshold and will be rejected otherwise. \par

To evaluate the impact of EER on user experience, we construct a combined metric called "effective latency", to represent the expected time a user will take to be authenticated, taking into account the possibility of the user not being recognized on his or her first attempt (false rejection rate). \par

To calculate effective latency, we make the assumption that the probability of a user being incorrectly rejected is equal to the false rejection rate, which enables us to use the expected value of a geometric distribution. The effective latency becomes the inference latency divided by the false reject rate, at a fixed false accept rate. In this work, we consider two false accept rates: 1\% and 10\%. A user desiring a more secure system would set a lower false accept rate; however, this would come at the expense of a greater inconvenience to the user, due to a greater average time taken to authenticate. \par

We recognize that the assumption of a geometric distribution may not always hold and that an already rejected user may be likely to be rejected again, despite being who they claim to be. Factors that contribute to an already-rejected user from being rejected again include the likelihood that samples taken in close temporal proximity are likely to be similar to each other. Hence, the cause behind a user being rejected in the first place (like poor lighting conditions that dimly illuminate the face or sickness that alters the user's voice) is likely to be present during the second round of authentication as well. Nevertheless, we use this quantity as it allows us to evaluate both EER and latency as a single metric. \par

\subsection{Models \& Datasets} \label{ssec:uist23:meth:mod}

We use the VoxCeleb2 and VoxCeleb1 audiovisual datasets for training and evaluation, respectively, since these datasets contain several thousands of unique identities and multiple samples for each identity \cite{Nagrani_2017, chung_voxceleb2_2018}. Additionally, the modalities contained for each identity (speaker and face) match the modalities captured by the ESP-EYE board, described in the next Section. Furthermore, the identities between VoxCeleb1 and VoxCeleb2 are non-overlapping, allowing us to train and evaluate our embeddings on completely different identities. By evaluating unseen identities, we mimic the generalizability of our authentication tool to new users which the model has not been trained upon. Both datasets are structured similarly, with several photos and audio samples corresponding to each identity. We use the dev split of VoxCeleb2 to train and the "vox-o" verification split of VoxCeleb1 to evaluate. The dev split of VoxCeleb2 contains face/voice samples and their corresponding identity label, consisting of over 1 million samples and 5994 unique identities. The "vox-o" split of VoxCeleb1 consists of 37720 pairs of face/voice samples drawn from the original 1251 identities present in VoxCeleb1 (although we randomly select  per training epoch). Half of the pairs belong to the same individual, while the other half correspond to different individuals. \par

Our high-level training script is based on a training script provided to us by the authors of \cite{Chen20VoxCelebMultimodal}. Additionally, the modified XNOR/fixed-point version of ResNet is based on Torchvision's open-source floating-point implementation \cite{torchvision_resnet}. \footnote{The complete
  code is available at https://github.com/<redacted for blind review>}

During inference, we remove the last fully connected layer, which generates the classification prediction during training. This truncated model outputs a 512x1 embedding, which (after normalization) represents the location of an identity on a 512-dimensional unit sphere. This embedding is only useful when compared with another embedding to produce a distance metric (Euclidean distance in our case). To generate a prediction, this distance value is compared to a threshold distance value chosen by the developer. Distance values below the threshold indicate the prediction that the identity of the two individuals is the same, while distance values above the threshold indicate that the two individuals are different.

During the evaluation phase of training, a threshold distance value is not chosen explicitly, but rather as the crossover value at which the false-accept and false-reject rates (FAR, FRR) intersect each other. The FAR/FAR value at this threshold (both are equal) is termed as the equal error rate (EER). By presenting many positive/negative pairs to the model at a certain state in training, we come up with a set of distance-label pairs. These distance-label pairs can be converted to smooth FAR-FRR curves by choosing thresholds at a fine level of granularity. For inference purposes, we choose the threshold value corresponding to a certain FAR value to afford a particular degree of security.

 In fusion models, we generate the embedding by simply concatenating the face embedding to the speaker embedding in a process referred to as feature concatenation \cite{radu_multimodal_2018}. Other possible methods of model fusion for future research include soft-attention fusion, compact bilinear pooling fusion, and gated multi-modal fusion. \cite{Chen20VoxCelebMultimodal}.

All model sizes are chosen such that parameter size is within 4MB, based on the available non-volatile memory on the ESP-EYE board. \par

% Unsure what figure to add, so commenting for now
% \todotbd{Insert figure here.}

\subsection{Deployment} \label{ssec:uist23:meth:depl}

We chose to deploy our multimodal model onto the ESP-EYE board because it fulfills the requirements for our system: low-cost, multi-core, vision/sound enabled. Some alternatives to the ESP-EYE include the ESP32-Cam, a camera-only microcontroller by Espressif, and a sensorless microcontroller (e.g. ESP8266) with an Arducam camera module added on. \cite{aguilar2019smart, kaur2021cloud} \par

For image acquisition, we capture an image from the camera and crop and resize it to 240x240 before passing it as input into the face verification model. From start to finish, this image preprocessing step takes approximately 120ms. For voice acquisition we capture a 3-second sample from the board, generate the mel-frequency cepstral coefficients corresponding to the sample, and pass the sample to the voice model \cite{Ali18MultimodalAuth}. Audio preprocessing latency is dominated by acquisition itself. In an ideal scenario, we would prefer methods with similar acquisition and preprocessing latency to balance the whole pipeline, but it is beyond the scope of this work. \par

\subsubsection{Latency estimation}
\label{ssec:uist23:disc:lat_est}

Additionally, we estimated the latency by computing the operation counts (opcount) required for each type of layer on the board (e.g. floating-point binary-weight convolution) and multiplying by the latency per opcount for each layer.

To run floating- and fixed-point quantized models on the ESP-EYE board, we use Espressif's ESP-DL inference library \cite{espressif_systems_esp-dl_nodate}, which contains optimized implementations of floating-point and fixed-point convolution and fully-connected layers.

For binarized models, we use the 3PXNet inference library \cite{Romaszkan20203pxnet} for on-device inference, with the following changes. Firstly, the single-bit functions of the original inference library are replaced with their corresponding multi-bit functions. Secondly, we implement additional functions, like element-wise addition and ReLU, which are not present in the original library, and are necessary for ResNet implementation.

% Secondly, we introduce an Im2Col version of the first ResNet layer that can handle pack sizes that are not multiples of 32, as in the input.

%For demonstration purposes, we interface with the user through LEDs and tactile buttons on the ESP-EYE. The authentication process is initiated with a push of the tactile button. During the recording process, a white light is displayed. Following image capture and audio recording, during the preprocessing and model inference phase, a yellow light is displayed. Finally, to present the results of the authentication, the LED glows red (rejected) or green (approved) for a second before the program resets and the LED turns back off. \par

\subsection{Cost Evaluation} \label{ssec:uist23:disc:bom}

The ESP-EYE \cite{EspEye23} board costs \$19.9 in retail \cite{DigiKey23}, and includes two dedicated memory ICs: 8MB PSRAM, and 4MB non-volatile Flash. While were not able to determine the price of the exact modules used on the ESP-EYE board, we have sourced prices of comparable ones from Digi-Key \cite{DigiKey23}, shown in Table \ref{tab:uist23:disc:bom}. A combined price of over \$4 comprises a significant part of the overall bill-of-materials (BOM). With a microphone price of roughly \$2, and a camera price of \$7, and the size-dependent memory and Flash prices shown below, we can estimate the cost of boards with the minimum required hardware for each model being considered. Using the results demonstrated in Section \ref{sec:uist23:res}, we show how this BOM can be reduced based on the desired application.

\begin{table}[htbp]\caption{Sample price of sensors (camera, microphone) and memory ICs: PSRAM (ISSI), and Flash (Winbond) \cite{DigiKey23}. All memories have quad SPI interfaces. Highest volume/lowest price avialable chosen. }

\begin{tabular}{c c} 
 \toprule
 Item & Cost [\$] \\ [0.5ex] 
 \midrule
 \multicolumn{2}{c}{Sensors} \\
 \midrule
 Camera & 7.6 \\ 
 Microphone & 1.56 \\
\midrule
 \multicolumn{2}{c}{Processors} \\
 \midrule
 ESP32-C3 (single-core) & 1.1 \\
 ESP32-S3 (dual-core) & 3.52 \\
\midrule
  \multicolumn{2}{c}{Memory} \\
  \midrule
 PSRAM 1MB & 2.07 \\
 PSRAM 2MB & 2.48 \\
 PSRAM 4MB & 2.81 \\ 
 PSRAM 8MB & 3.3 \\
 PSRAM 16MB & 3.88 \\ 
\midrule
 \multicolumn{2}{c}{Storage} \\
 \midrule
 Flash 1MB & 0.44 \\
 Flash 2MB & 0.57 \\
 Flash 4MB & 0.73 \\
 Flash 8MB & 0.92 \\
% Flash 16MB & 1.44 \\
 \bottomrule
\end{tabular}

\label{tab:uist23:disc:bom}
\end{table}

From the above information, we can infer some interesting potential trade-offs. For example, reducing the quantization level from fixed point to XNOR saves approximately \$1.5 dollars for ResNet18, which is enough to boost a face-only model to a multimodal model with both voice and face inputs. \par

Another tradeoff is between sensors and storage size. Generally, adding a sensor is more expensive than adding storage. As per Table \ref{tab:uist23:disc:bom}, the cost of expanding the model from audio-only to face+voice exceeds the cost of adding 8MB of Flash and 16MB of PSRAM, the maximum values needed to run most of the models we tested. The cost of converting a face-only model to a multimodal model is comparable to upgrading PSRAM from 1MB to 16MB and Flash from 1MB to 8MB. Thus, one could sacrifice the extra modality to support much more complex models. 64MB of Flash, for example, could support larger ResNets like ResNet152 in XNOR or fixed-point precision, and ResNet18 or ResNet34 in floating-point precision. However, as we will elaborate on in Section \ref{sec:uist23:res}, this does not account for another important factor - processing latency.

The combination of the above factors enables us to generate a search space of models for the use case of authentication. The resulting space contains models of varying quantization schemes, model architecture complexity, and level of modality, These factors in turn influence the cost of the resulting system, the performance that the model can achieve, and the latency of the model on the board. \par

%\todotbd{Add camera/microphone cost, add 1-2 ESP32 price points with different characteristics (e.g. single-core/double-core)}

\begin{comment}

\begin{table}[htbp]\caption{Sample price of memory ICs: PSRAM (ISSI), and Flash (Winbond) \cite{DigiKey23}. All memories have quad SPI interfaces. Highest volume/lowest price avialable chosen. }
\begin{tabular}{lrr}
\toprule
 & \multicolumn{1}{l}{PSRAM [\$]} & \multicolumn{1}{l}{Flash [\$]} \\ 
 \midrule
1MB & 2.07 & 0.44 \\ 
2MB & 2.48 & 0.57 \\ 
4MB & 2.81 & 0.73 \\ 
8MB & 3.3 & 0.92 \\ 
16MB & 3.88 & 1.44 \\ 
\bottomrule
\end{tabular}
\label{tab:uist23:disc:bom}
\end{table}

\end{comment}

%%%%%%%%%%%%%%%%%%%%%%%%%
% Results
%%%%%%%%%%%%%%%%%%%%%%%%%
\section{Results} \label{sec:uist23:res}

% todo: add following table to github
\begin{comment}
\begin{table*}[htbp]
\caption{Latency, cost, and accuracy of different hardware-model configurations. \hl{adjust as necessary} }
\label{tab:uist23:res:main}
\begin{tabular}{llllllll}

MCU & PSRAM & Flash & Camera & Microphone & Latency & Cost & Accuracy \\ 
ESP32 & 2MB & 2MB & \checkmark &  &  &  &  \\ 
 &  &  &  & \checkmark &  &  &  \\ 
 &  &  & \checkmark & \checkmark &  &  &  \\ 
 &  & 4MB & \checkmark &  &  &  &  \\ 
 &  &  &  & \checkmark &  &  &  \\ 
 &  &  & \checkmark & \checkmark &  &  &  \\ 
 & 4MB & 2MB & \checkmark &  &  &  &  \\ 
 &  &  &  & \checkmark &  &  &  \\ 
 &  &  & \checkmark & \checkmark &  &  &  \\ 
 &  & 4MB & \checkmark &  &  &  &  \\ 
 &  &  &  & \checkmark &  &  &  \\ 
 &  &  & \checkmark & \checkmark &  &  &  \\ 
\end{tabular}

\end{table*}
\end{comment}

%\subsection{Training Results} \label{ssec:uist23:results:train}

%\todors{Model training results, demonstrating how compression pareto curves with given bounds}

To generate EER results, we trained several versions of ResNet, varying in precision and the number of layers. We trained each model for at least 50 epochs and until the drop in evaluation EER remained less than 1\% over a 10-epoch period. For each trained model, we estimated the cost of a bare-minimum board required to run the model by adding and subtracting part costs from the base cost of the ESP-EYE to simulate inserting or removing hardware to/from the ESP-EYE. We also estimate the latency of the model through the process described in \ref{ssec:uist23:disc:lat_est}. The Pareto curves shown in Figures \ref{fig:uist23:res:pareto_size}, \ref{fig:uist23:res:pareto_lat}, and \ref{fig:uist23:res:pareto_cost} show the tradeoffs between these for parameter size, latency, and cost summarize our findings.

%TODO: I "eyeballed" the training trajectory to decide when to terminate a run, but there's no good way of writing this

\subsection{EER vs Parameter Size}

\begin{figure}[htbp]
    \makebox[\textwidth]{\includegraphics[width=0.7\paperwidth]{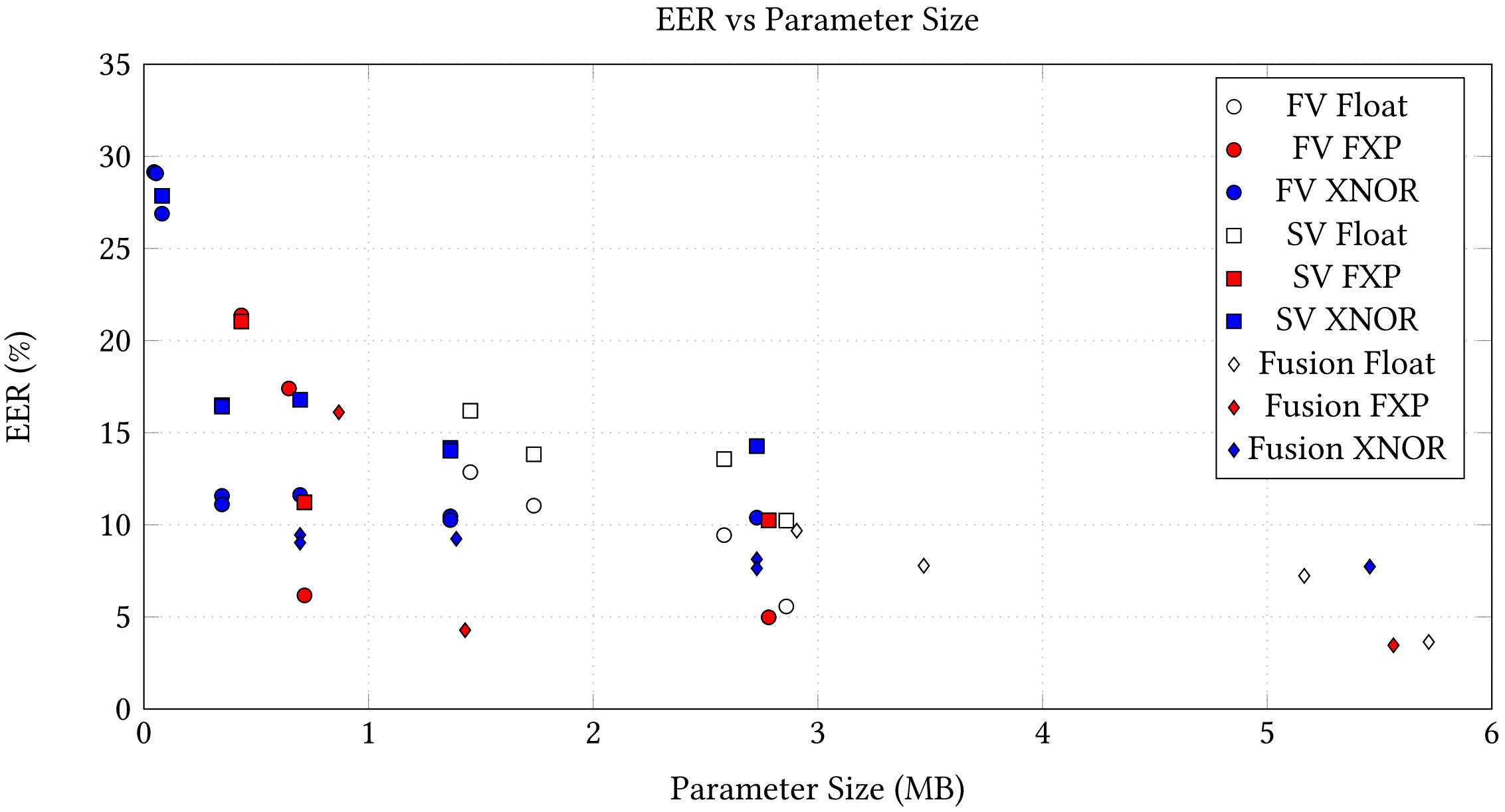}}
    \caption{For a given model, increasing the size of the model results in better performance as indicated by a lower EER.  However, changing the data representation of the model results in a different size/performance tradeoff, yielding a richer pareto front across more design points.}
    \label{fig:uist23:res:pareto_size}
\end{figure}

Figure \ref{fig:uist23:res:pareto_size} indicates that XNOR precision is best suited for models in the 0-750KB range, while fixed-point ones dominate for larger-sized models. Given XNOR models of a certain modality, some models have different error rates despite having the same model size because of differing activation bitwidth. For example, an XNOR 2/1 ResNet10 model will take double the runtime of an XNOR 1/1 ResNet10 model. However, a larger activation bitwidth leads to greater runtime, due to the greater number of computations needed per activation.

Fusion FXP models yield superior EER, as the best size-eer model belongs to a Fusion FXP model with a parameter size of <1.5MB and an EER of <5\%. However, this data point alone cannot be used to conclude that FXP models, in general, are better, as it ignores factors like latency and cost. This model has an estimated runtime latency of 2 minutes, which lies on the upper tail end of the distribution.

Another trend indicated by this graph is the diminishing returns associated with increasing model complexity. The face-only XNOR data points illustrate this pattern, for example. The increase in parameter size from 0.05MB to 0.35MB decreases EER by nearly 18 percentage points from 29.15\% down to 11.57\%. However, the much larger increase in parameter size from 0.35MB to 2.73MB minimally improves EER from 11.57\% to 10.39\%. This initial rapid rate of decrease followed by a much more gradual decrease suggests that for any given modality-precision combination, EER becomes saturated after a certain degree of model complexity.

Additionally, there is an upper limit associated with each modality. Initially, at small parameter sizes, face-only models perform the best, followed closely by speaker-only models. Eventually, however, voice saturates at a higher error rate than face-only and fusion models. This saturation leaves multi-modality as the only way to gain an additional improvement in EER.

\begin{comment}
\begin{figure}[htbp]
    \centering
    \includegraphics[width=9cm]{uist23/eer_vs_parameter_size.png}
    \caption{For a given model, increasing the size of the model results in better performance as indicated by a lower EER.  However, changing the data representation of the model results in a different size/performance tradeoff, yielding a richer pareto front across more design points.}
    \label{fig:uist23:res:pareto_size}
\end{figure}
\end{comment}

\subsection{EER vs Latency}

\begin{figure}[htbp]

    \makebox[\textwidth]{\includegraphics[width=0.7\paperwidth]{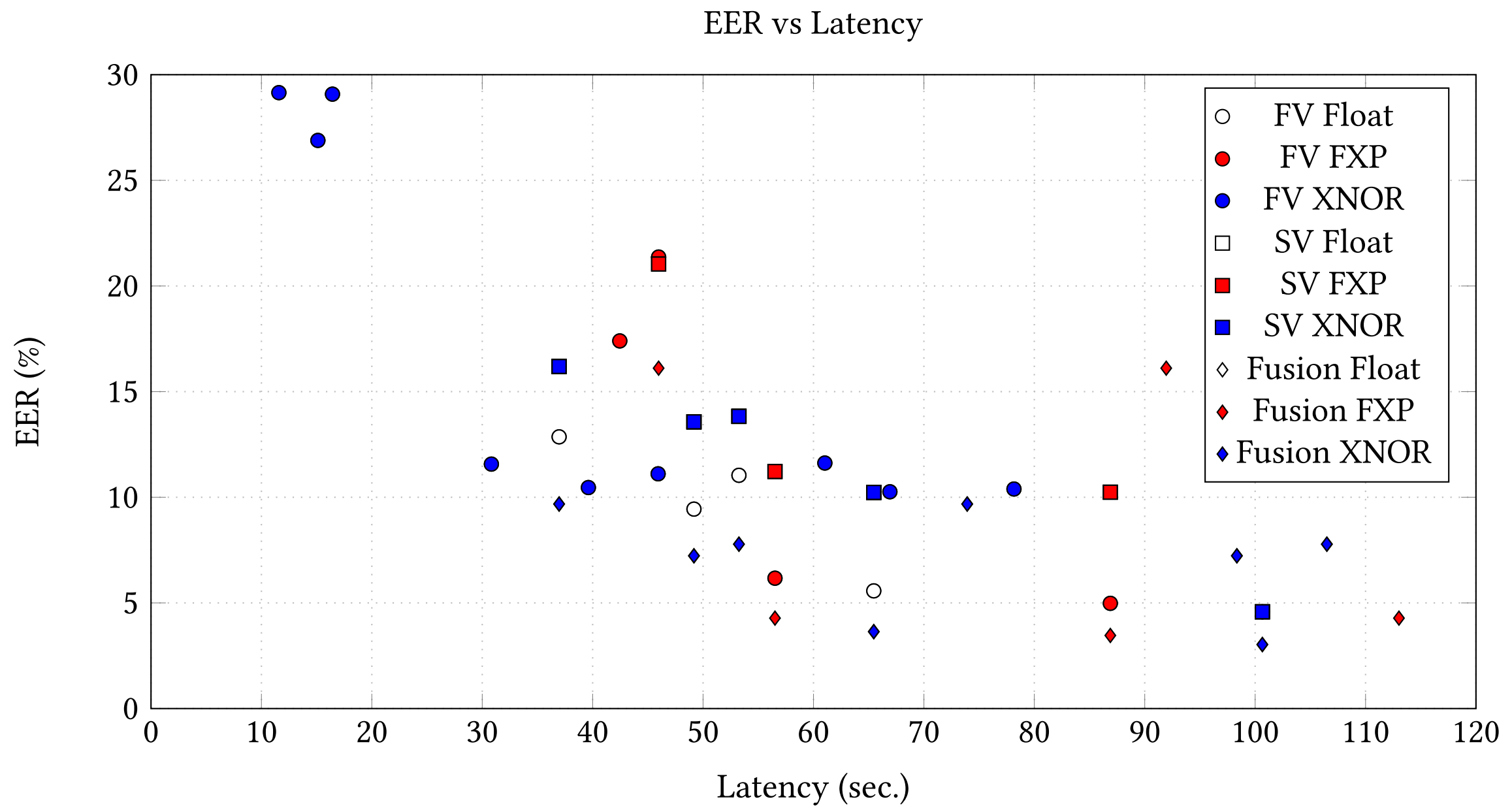}}

    \caption{Generally, XNOR models run much faster than floating- or fixed-point models of the same type. Within the XNOR regime, slightly more complex models yield a vast EER payoff. So, an improved error rate can be obtained with a more complex model, requiring a few more seconds of runtime.}
    \label{fig:uist23:res:pareto_lat}
\end{figure}

Figure \ref{fig:uist23:res:pareto_lat} displays the relationship between error rate and latency on the dual-core ESP32-S3 chip, excluding preprocessing time. For better interpretability, 10 outlier data points with >120 second latency have been omitted from the graph. These data points consist mostly of large floating-point models (e.g. floating-point fusion ResNet18).

Generally speaking, the majority of authentication models we trained are estimated to take between 30 and 120 seconds to execute (excluding input preprocessing time), with a few small XNOR models taking fewer than 30 seconds to run and a couple of large floating-point models taking more than 120 seconds to run (not shown on the graph). The most EER payoff can be seen in the sub-70 second range. For example, the best-achievable EER for a given latency steadily declines from around 11\% to just under 5\% when increasing the allowed latency from 30 to 60 seconds. Models that take longer than 70 seconds to run do not yield additional accuracy benefits, however. The likely explanation for this phenomenon is a saturation of model complexity for this use case, as was seen in the tradeoff between EER and Parameter Size.

Similar to the trend we saw when varying parameter size, as we increase latency, face-only models initially perform the best but are subsequently overtaken by fusion models in the > 30-second range. We also see a saturation of EER for each modality, for which increasing model complexity does not yield much additional benefit. Due to the time efficiency of XNOR and FXP model inference, these data points are Pareto superior to floating point data points, despite the accuracy degradation.

% As shown in this Pareto curve, though XNOR models do not achieve the best accuracy, they are the only precision that can be executed in a reasonable amount of time relative to their error rate. The majority of XNOR models shown in the figure are able to be executed in under 5 seconds. By comparison, the fastest fixed-point model takes 11.2 seconds to execute, which is an unreasonable amount of time for this application. Furthermore, if parallelism on the dual-core ESP-EYE is exploited, then the fusion model will generate its outputs in the same amount of time as a single-modal model. Notably, some floating point models with high latency perform worse than models with lower latency. This can be explained by the poor performance of ResNet18 compared to ResNet14, likely because the high complexity of ResNet18 leads to overfitting and poor generalization.

\subsection{EER vs Cost}

\begin{figure}[htbp]

    \makebox[\textwidth]{\includegraphics[width=0.7\paperwidth]{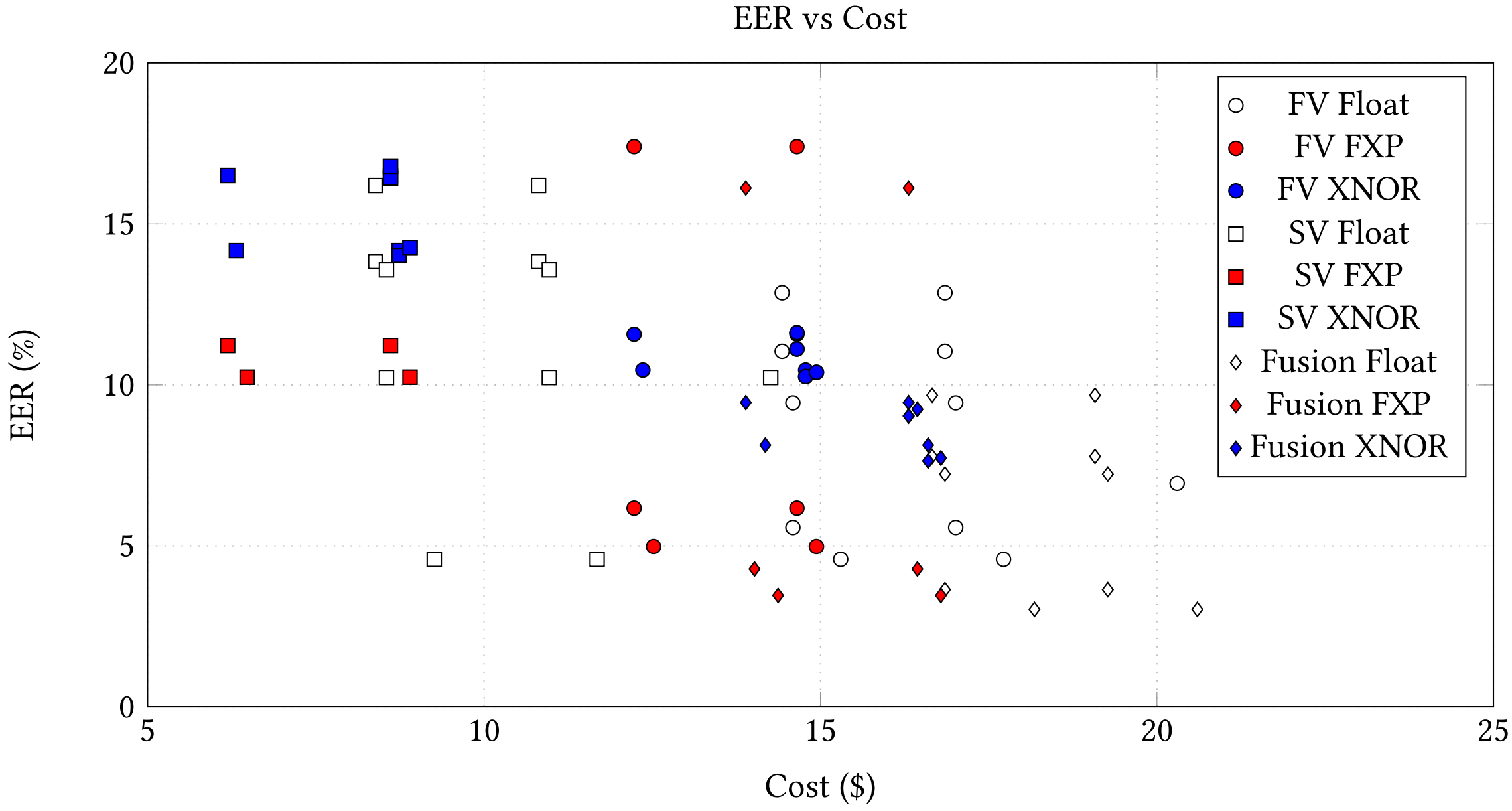}}

    \caption{Voice-only models are the cheapest, followed by face-only models, and followed by bimodal models with both face and voice recognition. However, the opposite trend is seen with respect to cost, with bimodal models performing the best and voice-only models performing the worst.}
    \label{fig:uist23:res:pareto_cost}
\end{figure}

Figure \ref{fig:uist23:res:pareto_cost} showcases three regimes: boards with cameras, boards with microphones, and boards with both devices. Naturally, the boards without cameras (SV - speaker verification) are the cheapest, though they suffer from a higher error rate, due to speaker verification's overall worse accuracy. Generally speaking, EER decreases with cost, though ResNet18 creates an exception to this trend, as in Figure \ref{fig:uist23:res:pareto_lat}.

Floating- and fixed-point models lie on the Pareto curve, with XNOR models trailing behind in terms of EER. Changing the precision of a model from FXP to XNOR causes a rise in EER while reducing cost. However, because Flash is relatively cheap, XNOR precision is not optimal from a cost-EER standpoint. Hence, a cost-centered analysis of the results leads to a different conclusion than a latency- or parameter size-centered analysis. As we saw in the analyses of Cost and Parameter Size versus EER, rather than XNOR/Face and FXP/Fusion models being Pareto-optimal, FXP/Voice and Float/Fusion models are optimal. Hence, given different optimization priorities, different choices of modality and precision become optimal.

\subsection{Cost vs Effective Latency}

\begin{figure}[htbp]

        \makebox[\textwidth]{\includegraphics[width=0.7\paperwidth]{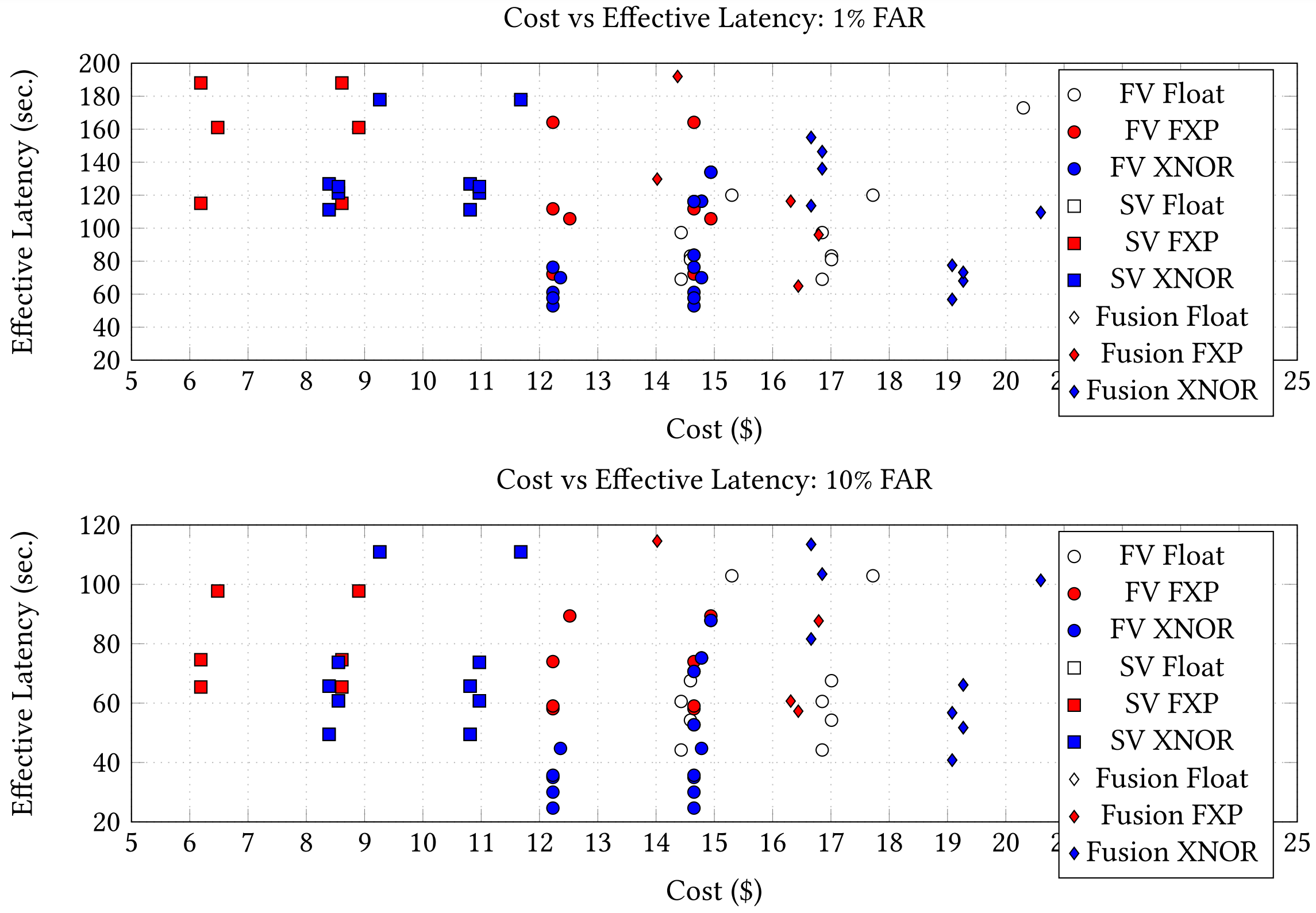}}

        \caption{The three graphs indicate the tradeoff between cost and effective latency in order of increasing false acceptance rates (FAR). As the FAR decreases, so does the effective latency due to false rejects that result in repeated authentication attempts. In general, speaker-only models have much higher effective latency compared to face or fusion models, despite their lower cost. Below the \$10 range, fixed-point is a better selection since it lies lower on the curve, while XNOR is a better option in the <\$10 range. the lowest effective latency can be obtained by selecting an XNOR or fixed-point model.}
        \label{fig:uist23:res:pareto_eff_lat}
    \end{figure}

Figure \ref{fig:uist23:res:pareto_eff_lat} displays the tradeoff between cost and effective latency at 2 different FAR levels: 1\% and 10\%. For models whose bare-minimum hardware costs more than \$10, XNOR tends to be the optimal choice of precision. In our case, 1\% FAR models will be unusable due to their high effective latency. Using a 5\% or 10\% FAR would be a more practical choice, despite the sacrifice in security.

% On average, our data indicate that an Int8 model is typically \hl{XX}\% faster than a Float model and an XNOR model is \hl{YY}\% faster than an Int8 model. However, when repeating the calculation with effective latency instead, the gap between Float and XNOR widens to \hl{ZZ}\%, and XNOR models outperform Int8 models by \hl{AA}\%.

\subsection{Analysis and Takeaways}

\subsubsection{Is it better to increase model complexity or add a modality?}

Our results indicate that the accuracy payoff of increasing model complexity is higher for simpler models but decreases as the models increase in complexity. For example, the EER reduction of a floating-point face-only model from ResNet-6 to ResNet-10 (~2x parameter count, \$0.15 cost increase) is over 7\% but the EER reduction from ResNet-10 to ResNet-14 (~4x parameter increase, \$0.70 cost increase) is a mere 1\%. So, for low-complexity models, it is generally better to add more layers to the model to boost accuracy, while more complex models may instead benefit from the addition of another modality. Overall, increasing the complexity of the model is a decision that yields diminishing returns with a growingly complex model.

The cost of adding another modality includes the cost of the additional sensor and Flash/PSRAM costs associated with the greater complexity of the model. In the Flash/PSRAM ranges we tested, the cost of doubling Flash/PSRAM size is under \$1, which is less than the cost of adding a microphone.

The error face-only models lag behind that of their multimodal counterparts by 1-4\% while speaker-only models trail behind by a larger 6-10\%. This error differential tends to be higher for less complex models. But while the EER-reduction benefit of going from a speaker-only to a multimodal model is considerable, the camera cost (\$7.6) is also nontrivial. Hence, if greater accuracy is desired, it is advisable to first increase the single-modality model's complexity before adding another modality, particularly for smaller-size models. 

Importantly, floating- and fixed-point precisions of the same architecture model are quite close in EER. Yet, fixed-point models consume a fourth of the parameter space and activation space and incur a slightly lower cost board than their floating-point counterparts. As a result of this tradeoff being extremely beneficial, a fixed-point model is almost always a better alternative compared to a floating-point model from a cost and latency point of view. This is consistent with established practices in the field of edge ML, which favor quantized datatypes \cite{Lai2018MlCmsis}. We opted to keep floating-point as a reference point, and to provide a more thorough analysis. Despite that, it is worth noting that floating-point models have their use - from the cost perspective they do provide some pareto-optimal solutions, e.g., in the $\sim\$10$ range.

Parallelizability is another important factor to consider when adding another modality to the board. For example, due to the dual-core processor on the ESP-EYE, or ESP32-S3, it is natural to run both face and speaker models, which cuts the overall latency in half compared to an approach with no concurrency. Concurrency is still possible, albeit more complicated, with a single-modality system deployed on a dual-core processor. For example, computation on the two branches of a convolutional block can be split between the two cores of a dual-core processor. However, for simplicity, we do not account for this fact in our latency estimates.

Depending on the use case, better strategies can be used to minimize user wait time. For example, an audio sample can be captured on the ESP-EYE while simultaneously processing the previously captured face data. This is not possible in our case since image and audio capture are simultaneous on the ESP-EYE, but this concept can be applied in other scenarios.

\subsubsection{How to choose the right precision?}

The impact of the precision of a model on the model's cost is minimal for two reasons: cheap cost of Flash, and similar PSRAM sizes across models.

Due to the cheap cost of off-chip Flash and its nonlinear scaling property, the additional expense of drastically improving the precision of a model is small. The cost difference between 1MB and 16MB of Flash is \$1, which is less than the cost of a microphone sensor. Hence, for a given precision, more complexity can be added to a model at a cheap cost. Yet, extreme binarization heavily reduces the expressive power of a model, leading to an increase in error rate of a few percentage points.

% If, however, the cost of Flash were to increase or scale in a linear fashion, the impact of increasing parameter size would become more noticable. In that scenario, XNOR models would lie close to or on the the EER-Cost pareto.

The second reason why precision minimally affects cost is the non-linear relationship between activation bitwidth and peak activation usage. Regardless of the model's precision, the precision of the input is usually floating point in order to avoid early loss of information. In our case, the input image size is always 0.57MB (224x224x3*4 bytes). The input image size serves as a lower limit on the PSRAM size on the board since the entire image must fit into PSRAM.

Typically, over the layers of a convolutional neural network (CNN), the number of channels increases, but the per-channel size decreases at a faster rate. As a result, activation size has a tendency to decrease in deeper layers. Thus, in CNNs like VGGNet or ResNets, the peak activation size usually occurs immediately after the first layer, when the number of channels increases from 3 to 32, for example. The consequence of this property is that peak activation memory is unaffected by adding or removing layers deep in the network, which is what is done when transforming a ResNet from ResNet10 to ResNet14.

In a floating-point ResNet model, the peak activation size is due to the output of the first layer, which is 1.53MB (112x112x32x4 bytes). So, a 2MB PSRAM will be used. The peak activation size within an XNOR or fixed-point network is less than the size of the input (0.57MB), however, so a 1MB PSRAM can be used.

\subsubsection{Balancing Error Rate and Latency}

From a cost-accuracy standpoint, it is generally better to choose a high-precision model than a low-precision model. However, precision is a significant factor when taking latency of the model into account. A model that takes too long to run may render the final product frustrating, inconvenient, or outright unusable.

So how should the accuracy drop of a low-precision, badly performing model be balanced with its latency? At a given price point, the user wishes to have low error rate and low latency at the same time. To help make this determination, we compute effective latency for each model, the expected time for a true user to be authenticated. As discussed in Section \ref{ssec:uist23:meth:train}, effective latency combines the performance of a model with its time efficiency into a single metric at a particular security tightness (fixed false accept rate) by indicating the average time a user would take to get authenticated.

Generally, in the \$10+ price range, XNOR models tend to lie along the cost-effective latency pareto, indicating that XNOR precision is an optimal choice. In the <\$10 price range, fixed-point tends to be a better precision choice due to the higher higher accuracy drop of applying XNOR quantization to small speaker models.

One exception to the generally optimal performance of XNOR models can be seen with the few isolated blue points that lie far away from the Pareto curve. This deviation can be accounted for by the fact that using XNOR quantization with very simple models leads to a drastic accuracy drop. For example, applying XNOR 2/1 quantization to a floating-point face-only ResNet6 model causes EER to rise by over 16 percentage points from 12.86\% with floating point to 29.15\%. By contrast EER increases by less than 4\% when the same quantization process is applied to a face-only ResNet18 model. So, smaller models are more sensitive to quantization, particularly XNOR quantization.

As expected, speaker-only models are the cheapest to run, as they lack a camera, followed by face-only models, and finally fusion models. But because speaker models tend to perform significantly worse than the other two categories, the effective latency of speaker-only models is penalized.

\subsection{List of Recommendations for ML-Hardware Codesign}

We distill the insights obtained to the following list of guidelines that a developer can use when balancing hardware and ML-model decisions.
\begin{itemize}
    \item \textit{Establish desired properties}. The Pareto-optimal precision/modality/complexity recommendations vary depending on whether the developer wishes to prioritize parameter size, latency, or model complexity. For example, XNOR-precision data points tend to be Pareto-optimal for the EER-latency and EER-parameter size tradeoffs, but not for the EER-cost tradeoff. The developer can analyze the tradeoff between model quality (in our case, effective latency) and cost to the developer (in our case, board cost).
    \item \textit{Choose an appropriate security level.} If greater security is desired, the FAR should be lower. However, lowering FAR increases FRR, which increases effective latency and decreases convenience. To maintain effective latency while tightening security, a costlier, more complex model-hardware combination may be necessary.
    \item \textit{Exploit parallelism opportunities.} The presence of a multi-core system opens a natural avenue to parallelizing the inference pipeline of a multimodal system. Particularly if the pipeline contains nonintersecting streams of execution for each modality, the workload can be naturally delegated across multiple cores.
    \item \textit{Experiment with quantization levels.} As our results indicate, FXP models sacrifice hardly any accuracy, while cutting down the parameter space by 4x and likewise reducing inference time. If greater accuracy degradation can be tolerated, low-bitwidth XNOR models can further reduce the required parameter space and heavily accelerate inference via the XNOR/popcount operators.
    \item \textit{Saturate model complexity before adding a modality.} The accuracy payoff of increasing a model's complexity is especially large for low- to medium-complexity models but reduces as the model's complexity increases. Hence, before another modality is added, particularly one corresponding to an expensive sensor, the developer should first ensure that the model's performance is not bottlenecked by a low-complexity model.
\end{itemize}

Note that the design space is non-smooth: depending on the choice of performance objectives, moving along a pareto curve may involve discrete non-adjacent transitions, e.g. as individual sensors or quantization schemes may be entirely replaced by others.  That is, allowing additional system cost from a pareto optimal solution may require hardware redesign (e.g. as a microphone is replaced by a camera), model retraining (e.g. with binarization instead of fixed point quantization), or both; otherwise, a resulting incremental advance may no longer be a pareto optimal design.

%\subsection{Runtime Results} \label{ssec:uist23:results:runtime}

%\todors{Showing that models can be successfully deployed and with reasonable runtime, plus integration with the device (camera/microphone)}

%Figure \ref{fig:uist23:res:runtime} shows runtime results of the entire pipeline on the ESP-EYE platform. \par

%\begin{figure}[htbp]
%    \centering
%    \missingfigure{ESP-EYE runtime results.}
%    \caption{Runtime results of proposed models on the ESP-EYE platform.}
%    \label{fig:uist23:res:runtime}
%\end{figure}

%%%%%%%%%%%%%%%%%%%%%%%%%
% Discussion
%%%%%%%%%%%%%%%%%%%%%%%%%
\section{Discussion} \label{sec:uist23:disc}

\subsection{Modality Choice} \label{ssec:uist23:disc:modal}

In this work we chose face and voice as modalities, due to their wide applicability and availability of camera and microphone sensors on the ESP-EYE board. However, upon analyzing the predictability, as indicated by accuracy results it may not be the best combination of modalities.

\begin{comment}

\begin{figure}[htbp]
    \centering
    \includegraphics[width=9cm]{uist23/pos_neg_hist.png}
    \caption{Embedding distances for mismatched face/speaker pairs. Generated by floating-point ResNet18 face model.}
    \label{fig:uist23:res:pos_neg_hist}
\end{figure}

\end{comment}

\begin{figure}[htbp]
\begin{comment}
\begin{tikzpicture}
\begin{axis}[
grid=major,
ymin=0,
xlabel=Distance,
ylabel=Frequency,
width=12cm,
height=8cm,
ybar
]
\addplot +[
red!60,
fill opacity=0.4,
hist={
    bins=15,
    % data min=120,
    % data max=230
    data min=0.4,
    data max=1.7
}   
] table [y index=0,col sep=comma] 
{chart_data/dist_histogram/negative.csv};
\addplot +[
blue!60,
fill opacity=0.4,
hist={
    bins=15,
    %data min=120,
    %data max=230
    data min=0.4,
    data max=1.7
}   
] table [y index=0] 
{chart_data/dist_histogram/positive.csv};
\legend{negative,positive}
\end{axis}
\end{tikzpicture}
\end{comment}
\makebox[\textwidth]{\includegraphics[width=0.7\paperwidth]{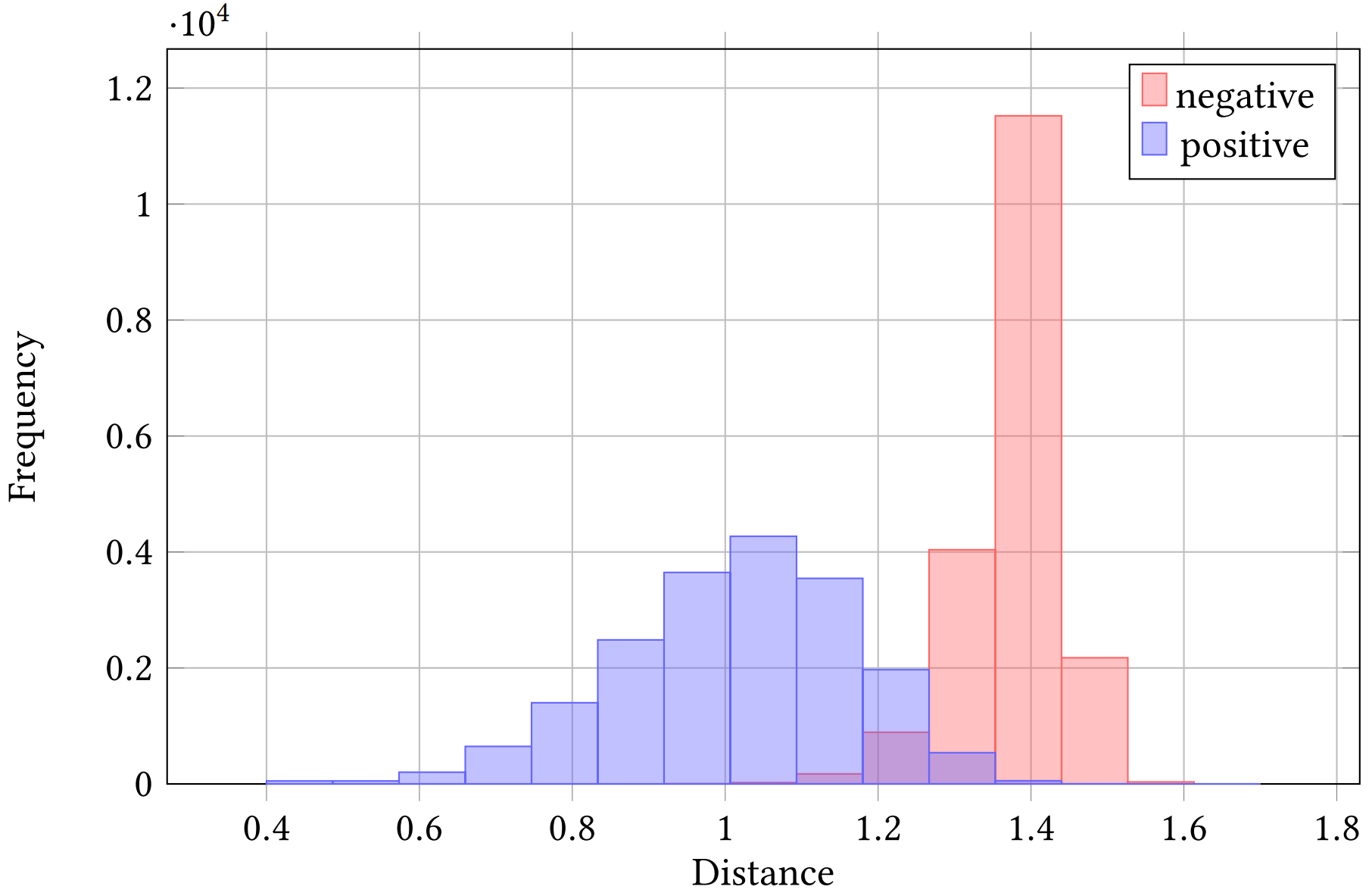}}
\caption{Embedding distances for mismatched face/speaker pairs. Generated by floating-point ResNet18 face model.}
\label{fig:uist23:res:pos_neg_hist}
\end{figure}

We start by analyzing how the closeness between two embeddings for positive and negative labels changes with modality. Figure \ref{fig:uist23:res:pos_neg_hist} shows the distribution of the distances between embeddings generated by a floating-point, ResNet18 face model. As expected, most of the time, two identities with the same label are on average closer to each other if they correspond to samples coming from the same identity. However, we noticed that the separation between the two distributions is less pronounced in the speaker model, suggesting that the speaker model alone is a worse predictor than the face model. \par

Ideally, given independent modalities, we would expect that combining the two models leads to an error rate equal to the product of the error rates of the constituent models. However, due to the correlation between face and speech, this assumption of independence does not hold true. As first discovered by \cite{Nagrani18CrossModalBiometric}, two identities with close face embeddings are likely to have close speaker embeddings and vice versa. As a result, while adding a second modality reduces the equal error rate of the model, the fusion model's error rate is higher than the product of the error rate of the models used to create the fusion model.

%seeing voices and hearing faces

\subsection{Architecture Family Choice}

One characteristic of ResNet models that led us to experiment with this architecture specifically is its modular nature, which allows us to add and remove blocks to change the number of parameters. By adding and removing the number of identity and convolution blocks in the model, it is straightforward to change the model's parameter count and expressive power. Though the number of layers in a standard ResNet models varies from 18 to over 150, we further downsize these models to as few as 6 layers. Another reason motivating our choice of ResNet models is that they have been proven to work well with the image inputs, making them an appropriate choice for both the image and audio spectrogram modalities \cite{Ding20AutoSpeech}. 

Another possible architecture choice was VGGNet, which includes the VGG-16 and VGG-19 architectures. Like ResNet, VGGNet has been demonstrated to work well with image inputs \cite{Ding20AutoSpeech}. However, with 138 million parameters, VGG16 exceeds the number of parameters in ResNet18 by roughly 12x, making this style of architecture too complex to experiment with. Even with 50\% sparsity and 1-bit quantization, the parameter count of VGG16 would be 8.63MB, which is too large to fit in the flash of most microcontroller boards, including the ESP-EYE.

On the smaller side, MobileNet is a lightweight architecture family with a parameter count ranging from 1.6 million to 5.4 million parameters. Although the relative size of MobileNet is appropriate for deployment onto the ESP-EYE, its lack of repeated blocks comparatively difficult to vary its complexity while preserving the overall architecture. %Furthermore, the presence of depthwise and separable convolutions in the MobileNet architecture would introduce further implementation complexity, which we sought to avoid for the purpose of simplicity.

The choice of model architecture and corresponding dataflow also has an impact on peak activation memory, and thus the required PSRAM size. As pointed out earlier, ResNet activation memory size is dominated by the input layer, and so quantization and smaller networks have negligible impact on PSRAM size and cost. This dependence can change, for example,  for a VGG-like architecture where several dense layers can dominate activation memory size making quantization impact on cost more pronounced. 
\subsection{Automated Exploration} \label{ssec:uist23:disc:auto}

The work presented in this manuscript marks an important first step in cost-driven edge ML deployment. It establishes relationships between different compression techniques for different deployment scenarios. However, it is inherently limited by the choice of application, devices, sensors, models, etc. As an end goal, we believe there is a need for an automated tool that could take into account all of the above, and provide a set of ready-to-deploy candidate solutions. Such a tool would serve as an ultimate enabler of accessible ML, as it would be able to quickly answer the question of what is feasible given a certain budget. While this automated framework is a subject for future work, we want to highlight that the methods and data presented here form a solid foundation for it. \par

\section{Conclusion}
In this work, we have demonstrated the tradeoffs between hardware design parameters---including processor provisioning, memory sizing, and sensor selection---and software design parameters---including model architecture selection, weight and activation quantization, and sensing modality--for low cost, accessible edge devices tasked with machine learning inference.  We used these results in a case study to optimize a multimodal edge authentication system in a way that accounts for its performance, cost, and user-experience. Through a search of model architectures, quantization schemes, and modality choices, we determined that XNOR/FXP authentication models tend to be the most latency-efficient, while FXP/Floating-Point models are the most cost-efficient. We also demonstrate that despite the slight correlation between face and voice modalities, the combination of the two into a fusion model yields significant model improvement at minimal additional runtime overhead when multi-threaded processing is employed.

\begin{acks}
\end{acks}

\bibliographystyle{ACM-Reference-Format}
%\bibliography{references_zotero}
\bibliography{references_ravit}

%%
%\appendix

\end{document}